\documentclass{article} 
\usepackage{iclr2026_conference,times}
\usepackage{booktabs}

\usepackage{amsmath,amsfonts,bm}

\def\eqref#1{equation~\ref{#1}}

\def\1{\bm{1}}

\DeclareMathAlphabet{\mathsfit}{\encodingdefault}{\sfdefault}{m}{sl}
\SetMathAlphabet{\mathsfit}{bold}{\encodingdefault}{\sfdefault}{bx}{n}

\usepackage{hyperref}
\usepackage{url}

\title{Probabilistic Dreaming for World Models}

\author{Gavin Wong \\
Yale University\\
New Haven, CT\\
\texttt{gavin.wong@yale.edu} \\
}

\iclrfinalcopy 

\makeatletter
\def\@noticestring{%
  Presented at the ICLR 2026 Workshop on <Workshop Name>.%
}
\makeatother

\begin{document}

\maketitle

\begin{abstract}
``Dreaming" enables agents to learn from imagined experiences, enabling more robust and sample-efficient learning of world models. In this work, we consider innovations to the state-of-the-art \textit{Dreamer} model using probabilistic methods that enable: (1) the parallel exploration of many latent states; and (2) maintaining distinct hypotheses for mutually exclusive futures while retaining the desirable gradient properties of continuous latents. Evaluating on the MPE SimpleTag domain, our method outperforms standard Dreamer with a 4.5\% score improvement and 28\% lower variance in episode returns. We also discuss limitations and directions for future work, including how optimal hyperparameters (e.g. particle count $K$) scale with environmental complexity, and methods to capture epistemic uncertainty in world models. 
\end{abstract}

\section{Introduction}

In reinforcement learning (RL), ``dreaming" refers to the process where an agent learns by imagining future trajectories using a learned world model, rather than solely interacting with the physical environment \citep{ha2018world, hafner2019planet}. This paradigm was popularized by Hafner et al. (2020) with \textit{Dreamer}, which achieved state-of-the-art performance across diverse domains (Atari, Minecraft, robot locomotion) while requiring orders of magnitude less training examples and minimal hyperparameter tuning \citet{hafner2020dreamer}. This mode of `latent imagination' has since become the standard in model-based RL. 

In this paper, we consider a few innovations to Dreamer's latent imagination process by integrating probabilistic methods. These innovations were motivated by the observation that:
\begin{enumerate}
    \item Despite learning a full distribution of latent states, Dreamer only samples a single state to roll-out a single imagined trajectory, potentially limiting the agent's ability to explore the full breadth of causes during training;
    \item While recent iterations of Dreamer (v3/v4) have moved to discrete categorical latents to handle multimodality \citep{hafner2021dreamerv2, hafner2023mastering}, continuous Gaussian latents remain desirable for their smoother gradient properties and dense representation. Yet, standard unimodal Gaussians can bias the model towards a non-existent mean when facing distinct alternatives -- such as averaging ``left" and ``right" paths into an impossible ``middle" path. 
\end{enumerate}

We therefore explore 3 possible improvements to the latent imagination process: 

\begin{enumerate}
    \item \textbf{To resolve the multimodal averaging problem:} We use a particle filter to represent the latent distribution, allowing the model to maintain distinct, competing hypotheses about the future (e.g. separate particles for 'left' and 'right') while retaining continuous latents. 
    \item \textbf{To address single roll-outs:} We perform parallel roll-outs per particle, and employ a latent beam search allowing each particle to branch out into multiple actions per time-step. 
    \item \textbf{To maintain computational tractability:} We prune trajectories based on the ``free energy" principle, keeping only the trajectories with highest predicted score and information gain.
\end{enumerate}

Below, we implement these innovations and conduct a preliminary evaluation on a simple predator-prey game (MPE SimpleTag; \citealp{lowe2017maddpg, terry2021pettingzoo}). We intend for this work to be an initial proof of concept, showing how probabilistic methods could be used to improve model-based RL, while also highlighting insufficiencies in our existing method to direct further work. 

\section{Methods}

\subsection{Game Domain: MPE SimpleTag}

We evaluate all proposed models on the MPE SimpleTag domain, a simple game where the agent must evade capture from three predators. We design the predator behaviour to be intrinsically multi-modal: when a predator is within some pre-defined radius of the agent, it stochastically switches between its ``CHASE" and ``INTERCEPT" strategies, creating a branching state space where the predator's future position is bimodal. This relatively simple set-up allows us to isolate and compare the models' ability to maintain a belief over discrete, mutually exclusive strategies. 

\subsection{Model Implementation}

We build our backbone (`BaseDreamer') upon the Dreamer-v3 architecture, except we replace the categorical latent distribution with a Gaussian (as in Dreamer-v1/v2) to test our hypothesis regarding continuous latent gradients. All other components are kept the same: its world model learns a deterministic hidden state $h_t$ and stochastic latent state $z_t$, and consists of a Recurrent State Space Model (RSSM), a posterior encoder, a prior, a decoder, and prediction heads for reward and episode continuation. The posterior encoder is trained to predict a latent distribution $z_t$ given the current game observation $x_t$ and hidden state $h_t$, and the prior is trained to match the posterior without using observations from the environment, effectively allowing the model to `dream' about future trajectories without interacting with the environment. 

We then implement `ProbDreamer' by modifying BaseDreamer with the following:

\textbf{Particle Filter.} During imagination, instead of sampling a single latent state at each time-step, we maintain a set of $K$ particles $\{h^{k}_{t}, z^{k}_{t}\}^{K}_{k=1}$ that track the latent distribution given by the prior, giving us $K$ parallel dreams per training step. While each particle transition $p_\theta(z_{t+1}|h_{t+1})$ remains Gaussian, our belief over latent states is now an empirical distribution over particles, and after several steps of stochastic propagation and resampling can approximate complex, multi-modal beliefs. 

\textbf{Latent Beam Search.} Then, to propagate each particle, we explicitly branch each particle into $N$ candidate actions \citep{schrittwieser2020muzero}, sampled from the policy $\pi_\theta(a|h^k_t, z^k_t)$. This gives us $K*N$ branches, each of which we propagate using the world model. 

\textbf{Minimizing Free Energy.} \citep{friston2010free} Since we do not have access to real observations during the dreaming process, we cannot use standard MLE to prune particles based on which best explain the new observations. Thus, we score branches using both their predicted reward and the epistemic uncertainty of the world model in that latent state. Predicted value is given by the critic $V_\phi(h_t^k, z^k_t)$, while epistemic uncertainty is approximated by the disagreement in an ensemble of prior models \citep{osband2016deep, chua2018deep, sekar2020plan2explore}. To balance between exploiting high-reward trajectories and exploring novel situations with high uncertainty, we minimize (negative) free energy by maximizing its constituent components: 

\begin{center}
$F^k_t=V_\phi(h_t^k, z^k_t)+\beta \cdot \sigma_{ens}^2$
\end{center}

where $V_\phi(h_t^k, z^k_t)$ = critic predicted reward, $\sigma_{ens}^2$ = ensemble variance, $\beta$ = scaling factor. 

\subsection{Training Procedure}

We apply standard actor-critic learning in a loop that alternates between collecting experience and latent training. In each iteration, the agent collects $10^3$ real environmental steps for its replay buffer, followed by $2 \cdot 10^4$ steps of latent imagination where the policy is optimized against the world model. 

We retain the three standard loss functions used in Hafner et al. (2023) \citep{hafner2023mastering}:

\begin{enumerate}
    \item \textbf{Reconstruction loss.} Ensures the latent state $z_t$ is informative enough to reconstruct the observation $x_t$.
    \item \textbf{Dynamics loss.} Minimizes the KL-divergence between the prior (imagination) and the posterior (observation-grounded) latent distributions.
    \item \textbf{Representation loss.} Regularizes the posterior towards the prior to prevent overfitting and keep it close to what the prior can predict. 
\end{enumerate}

\subsection{Experimental Set-Up}

\textbf{Hyperparameter Search.} To ensure a rigorous comparison, we conducted an extensive Bayesian Optimization (BO) sweep to identify optimal hyperparameters for both the baseline and probabilistic models. We primarily tuned the number of particles ($K$), latent beams ($N$), imagination horizon ($T$), and learning rates. Architectural hyperparameters (e.g., hidden state size, RSSM capacity) were held constant to ensure that performance differences arise solely from the imagination mechanism rather than representational capacity. The sweep comprised 150 runs across 6 NVIDIA GPUs.

\textbf{Model Comparison.} From the sweep, we selected the 6 best configurations as 'finalists' to evaluate on 100 fixed test episodes across 5 random seeds. These 6 finalist models belong to 3 model classes:

\begin{itemize}
    \item \textbf{Baseline Dreamer:} The control model, with $K=1, N=1, T \in \{10, 16\}.$ 
    \item \textbf{``Lite" ProbDreamer:} A lightweight probabilistic variant with $K \in \{2, 4\}$ and no latent beam search ($N=1$). 
    \item \textbf{``Full" ProbDreamer:} The complete model 
    ($K \in \{4,8\}$) utilizing latent beam search \\ ($N \in \{2, 4\}$) and free energy pruning.
\end{itemize}

\section{Results}

\subsection{Probabilistic Latents Improve Performance and Robustness}

\begin{table}[h]
\centering
\label{tab:results}
\begin{tabular}{l c c c c}
\toprule
\textbf{Model} & \textbf{\textit{K}} & \textbf{\textit{N}} & \textbf{\textit{T}} & \textbf{Performance} \\
& & & & \small{(mean $\pm$ std over 5 seeds)} \\
\midrule
BaseDreamer 1 & 1 & 1 & 16 & $-9.21 \; (\pm 0.80)$ \\
BaseDreamer 2 & 1 & 1 & 10 & $-9.74 \; (\pm 0.79)$ \\
\textbf{ProbDreamer Lite 1} & \textbf{2} & \textbf{1} & \textbf{10} & $\mathbf{-8.79 \; (\pm 0.68)}$ \\
ProbDreamer Lite 2 & 4 & 1 & 22 & $-9.43 \; (\pm 1.57)$ \\
ProbDreamer Full 1 & 2 & 4 & 10 & $-53.78 \; (\pm 12.14)$ \\
ProbDreamer Full 2 & 8 & 1 & 22 & $-26.84 \; (\pm 23.03)$ \\
\bottomrule
\end{tabular}
\caption{Comparison of model performance (0 is perfect performance). Lite ProbDreamer 1 achieves best performance. $K$ = number of particles, $N$ = latent beams, $T$ = imagination horizon.}
\end{table}

``Lite" ProbDreamer ($K=2, N=1$) consistently outperforms BaseDreamer in 4 out of 5 seeds, averaging a 4.5\% score improvement. It also learns a more robust policy that leads to 28\% lower variance in episode returns. This result validates our hypothesis that representing the latent distribution as a particle filter allows the agent to more flexibly maintain competing hypotheses, such as the predator's distinct ``Chase" and ``Intercept" strategies. Indeed, by analyzing the gameplay footage, we noticed that the main difference in agent behaviour was that ProbDreamer is able to react quickly to changes in predator strategies, while BaseDreamer tends to freeze momentarily -- a telltale sign of the Gaussian bias collapsing mutually exclusive futures into a single, paralyzed mean.

\subsection{Challenges of Active Latent Pruning}

However, while the particle filter proved effective for robustness, we observed sharp performance degradation when introducing latent beam search and high particle counts, as seen in the ``Full" model results. We attribute this failure to three specific challenges:

\textbf{Particle Saturation.} An interesting result is how performance improves from $K=1$ to $K=2$, but degrades as we introduce more particles. One hypothesis for why $K=2$ is optimal could be that the tag predators have 2 distinct strategies (``Chase" and ``Intercept"), which require 2 particles to model; any larger $K$ and the model begins fitting noise. To test this, we plan to perform further work on how the optimal $K$ changes with the number of distinct modes or strategies across different domains. If true, this would imply particle count $K$ must be carefully selected, possibly using domain knowledge. 

\textbf{Ineffective Pruning Mechanism.} The poor performance of latent beam search likely points to poor selection of trajectories to prioritize during imagination. Our pruning mechanism relies on a value function $V(z)$ to select the best branches -- yet, during the imagination phase, there are no ground-truth observations to correct the model. This means when the critic is noisy during early training, it may assign falsely high values to unrealistic trajectories, which the model then preferentially selects for during imagination, leading to noisy training and failure to converge. We hence suggest that further work be cautious of methods involving pruning imagined trajectories solely based on a learned value function, since there is no ground-truth observation available to correct the model.

\textbf{Ensemble Collapse.} To balance exploitation with exploration, our scoring function included a curiosity term based on the disagreement of an ensemble of prior models ($\sigma_{ens}^2$). However, in practice, we found that the ensemble members rapidly collapsed to near-identical predictions, and an ablation study showed minimal difference when the curiosity metric was removed. In follow-on work, we suggest exploring more robust methods to estimate epistemic uncertainty, including: (1) explicitly diversifying the ensemble with separate optimizers and different episode replays; (2) using richer Bayesian approximations of uncertainty like Monte-Carlo dropout \citep{gal2016dropout}; or (3) also using disagreement in predicted rewards or observations, rather than just latent-state predictions. 

\section{Discussion and Conclusion}

In this work, we proposed 3 innovations to the latent imagination process popularized by \textit{Dreamer} using probabilistic methods. By evaluating on the MPE SimpleTag domain, where an agent learns to evade several predators that stochastically switch strategies, we show that probabilistically sampling multiple latent states as ``particles" leads to a consistent improvement in performance over methods that sample a single latent. Furthermore, the agent learns a significantly more robust policy. We suggest that this particle filtering method is an ideal way to: (1) enable parallel roll-outs of latents to explore the full breadth of causes; and (2) resolve multi-modal ambiguity while retaining the desirable gradient properties of continuous latents. We believe these positive results warrant further exploration into non-parametric world models. 

We also identify two primary limitations that point toward fruitful directions for future research. First, the relative simplicity of the fully observable MPE domain meant that a bimodal particle filter ($K=2$) was sufficient to capture the relevant predator strategies. This could have saturated the performance benefits of our probabilistic approach and prevented gains in sample efficiency. A meaningful follow-up would evaluate \textit{ProbDreamer} on partially observable and chaotic environments, to rigorously test how the optimal particle count ($K$) scales with environmental complexity. 

Second, we found that the primary bottleneck for active latent imagination is the lack of ground-truth observations to correct the model's ``dreams". Our attempt to prune trajectories using a ``free energy" objective, while well-motivated, struggled as a noisy value function biased the agent towards optimistic hallucinations, while the ensemble of prior models collapsed to near-identical predictions. Nonetheless, we remain convinced that development architectures that intrinsically capture epistemic uncertainty is key to unlocking the full potential of model-based RL. Solving this challenge would not only enable agents to autonomously balance exploration and exploitation (instead of relying on some fixed regularizer), but also bring us closer to algorithms that mirror the active, curiosity-driven learning observed in human cognition. 

\newpage 

\bibliography{iclr2026_conference}
\bibliographystyle{iclr2026_conference}

\end{document}